\documentclass{article}

\usepackage{PRIMEarxiv}

\usepackage[utf8]{inputenc} 
\usepackage[T1]{fontenc}    
\usepackage{url}            
\usepackage{booktabs}       
\usepackage{amsfonts}       
\usepackage{nicefrac}       
\usepackage{microtype}      
\usepackage{lipsum}
\usepackage{fancyhdr}       
\usepackage{graphicx}       
\graphicspath{{media/}}     

\usepackage{graphicx}
\usepackage{amsmath} 
\usepackage[misc,geometry]{ifsym}
\usepackage[bookmarksnumbered,bookmarksdepth=3,colorlinks=true,allcolors=blue]{hyperref}
\usepackage{multicol}

\usepackage[labelsep=period]{caption}
\usepackage[symbol]{footmisc}

\usepackage{etoolbox}
\apptocmd{\sloppy}{\hbadness 10000\relax}{}{}

\title{Physics-Informed and Hybrid Machine Learning in Additive Manufacturing: Application to Fused Filament Fabrication
\thanks{\textit{\underline{Citation}}: 
\textbf{Kapusuzoglu, B., Mahadevan, S. Physics-Informed and Hybrid Machine Learning in Additive Manufacturing: Application to Fused Filament Fabrication. \textit{JOM} 72, 4695--4705 (2020). DOI: \href{https://doi.org/10.1007/s11837-020-04438-4}{ 10.1007/s11837-020-04438-4}
}} 
}

\author{
  Berkcan Kapusuzoglu \\
  Department of Civil and Environmental Engineering \\
  Vanderbilt University \\
  Nashville, TN 37235, USA \\
  \texttt{berkcan.kapusuzoglu@vanderbilt.edu} \\
  \And
  Sankaran Mahadevan \\
  Department of Civil and Environmental Engineering \\
  Vanderbilt University \\
  Nashville, TN 37235, USA \\
  \texttt{sankaran.mahadevan@vanderbilt.edu} \\
}

\begin{document}
\maketitle

\begin{abstract}
This paper investigates several physics-informed and hybrid machine learning strategies that incorporate physics knowledge in experimental data-driven deep learning models for predicting the bond quality and porosity of fused filament fabrication (FFF) parts. Three types of strategies are explored to incorporate physics constraints and multi-physics FFF simulation results into a deep neural network (DNN), thus ensuring consistency with physical laws: (1) incorporate physics constraints within the loss function of the DNN, (2) use physics model outputs as additional inputs to the DNN model, and (3) pre-train a DNN model with physics model input-output and then update it with experimental data. These strategies help to enforce a physically consistent relationship between bond quality and tensile strength, thus making porosity predictions physically meaningful. Eight different combinations of the above strategies are investigated. The results show how the combination of multiple strategies produces accurate machine learning models even with limited experimental data.

\end{abstract}

\section{Introduction}\label{sec:intro}
Achieving the desired material properties and product quality in additive manufacturing (AM) processes has been studied using trial-and-error experiments as well as process models (either physics-based or machine learning (ML) models). In a trial-and-error approach, the AM process is repeated multiple times with different process parameter combinations to achieve the desired microstructure and properties of the manufactured parts; this is expensive and time-consuming. Moreover, this trial-and-error approach needs to be implemented every time a new design needs to be manufactured. Therefore, in recent years, research efforts have focused on model-based methods for optimizing the AM process parameters.

Several physics-based models have been developed depending on the AM process category and the quantity of interest (QoI)~\cite{debroy2017building}. Costa et al.~\cite{costa2017estimation} proposed an analytical solution for transient heat transfer during the printing process in fused filament fabrication (FFF). Different models have been proposed in the literature to study polymer sintering~\cite{pokluda1997modification,frenkel1945viscous,hopper1984coalescence,garzon2020design}. Many of these models are parametric representations of complex physical processes based on various approximations. The parameters of such physics-based models, as well as the model errors need to be calibrated for each AM process using available observation data to reduce the uncertainty in the model predictions~\cite{ling2014selection,nath2019uncertainty,berkcan}. Due to the complex physics of the AM process, a different model is needed for each sub-stage or phenomenon in the manufacturing process in order to accurately predict the QoI. Further, physics models with reasonable fidelity and accuracy require tremendous computational effort; as a result, the use of physics-based modeling in AM of realistic products has been challenging and limited.

Recently, several studies have used the available AM experimental data to build \emph{black-box} ML models. In addition, with increased computing power, deep learning has become a prominent tool for solving classification and regression problems. In the context of AM, Khanzadeh et al.~\cite{khanzadeh2018porosity} compared supervised machine learning approaches to classify melt pools to predict porosity. Artificial neural network has been used to predict the geometry of a single bead in wire and arc additive manufacturing from the wire-feed rate and travel speed~\cite{ding2016bead}. Kwon et al.~\cite{kwon2020convolutional} investigated the convolutional neural network (CNN) to predict laser power from melt pool images. Zhang et al.~\cite{zhang2019process} used a CNN model to perform in-process porosity monitoring of laser-based AM processes. 

Physics-based models do not require large amounts of data, but are generally limited by their computational complexity or incomplete physics. In contrast, ML models appear promising for complex systems that are not fully understood or represented with simplified relationships, given adequate quality and quantity of data. However, ML models represent the complex physics without taking into account any physical laws and thus can produce results that are inconsistent with physical laws. Karpatne et al.~\cite{karpatne2017physics} proposed the combined use of physics-based and ML models to achieve more accurate and physically consistent predictions by leveraging the advantages of each method. In order to make ML models consistent with physical laws, Karpatne et al.~{\cite{karpatne2017physics}} incorporated physical constraints into the loss function of ML models. Another method that is also implemented by Karpatne et al.~{\cite{karpatne2017physics}} to combine physics-based and ML models is to use the physics-based model outputs as additional inputs in an ML model along with other inputs. Jia et al.~{\cite{jia2020physics}} used synthetic data generated by executing physics-based models for multiple input combinations to pre-train the ML model in order to leverage the knowledge embedded in physics-based models. These general ideas have been explored in multiple engineering applications, such as geoscience, fluid dynamics, and thermodynamics, and this paper investigates these ideas for additive manufacturing.

The goal of this paper is to enhance the experimental data-driven ML models for AM by incorporating the physics knowledge, thus helping the ML model to produce more accurate and physically meaningful results. Three different strategies are explored for this purpose: (1) incorporate physics constraints within the loss function of the deep neural network (DNN), (2) use physics model outputs as additional inputs to the DNN model, and (3) pre-train a DNN model with physics model input-output and then update it with experimental data. Eight different combinations of the above three strategies are explored, and their performance in porosity and bond quality prediction of the FFF-produced parts is examined.

The strategies investigated in this paper help to better capture the physics of the AM process by leveraging physical laws while improving the generalization performance of data-driven AM models. Compared to previous studies, the proposed strategies address two different physical QoIs (neck diameter and porosity) and use multiple physics-based loss functions to enhance the ML model. The main contributions of this work are as follows:

\begin{enumerate}
    \item Several physics-informed and hybrid machine learning models are developed for porosity prediction of FFF parts using physics constraints, physics-based model, and experimental data, using the three strategies mentioned above.
    \item An enhanced physics-based model is developed to account for realistic filament geometry and the change of geometry during the printing process.
    \item The proposed models are trained and evaluated using laboratory experiments where FFF parts are printed with varying input conditions, and data is collected to measure the quality characteristics (bond quality, porosity) of the parts.
\end{enumerate}

The remainder of the paper is organized as follows. The proposed methodology is presented in Section~\ref{sec:method}, followed by the implementation of the methodology for a FFF process in Section~\ref{sec:results}. Concluding remarks are provided in Section~\ref{sec:conclusion}.

\section{Methodology}\label{sec:method}
This section develops the proposed physics-informed machine learning (PIML) approaches to predict the bond formation and mesostructure in FFF-produced parts. The methodology is applicable to any AM process with corresponding data and physical laws for the prediction QoI. The three components of the methodology are: (a) Physics-based models, (b) Experiments, and (c) Construction of PIML models.

\subsection{Physics-based models}\label{sec:method1}
This subsection describes the two coupled multi-physics models (thermal model and polymer sintering model) that predict the bond formation between adjacent filaments and the mesostructure of the printed part. The porosity and bond quality of an FFF part is dependent on the temperature history of filaments. Thus, it is important to predict the temperature evolution of filaments to estimate the final mesostructure of the printed part. The thermal model, based on the work by Costa et al.~{\cite{costa2017estimation}}, is used to predict the temperature evolution of filaments considering the material properties, part geometry, and process parameters. The output of the heat transfer model (temperature) is input to the sintering model to predict the porosity and bond quality. A new method is then developed, which considers realistic filament geometry, and allows the filament geometry to change during the printing process, to compute the rate of polymer sintering and the final mesostructure of the printed part using the predicted temperature evolution of each filaments. Thus the mapping from input to output is a multi-physics model, i.e., models of two physical phenomena (heat transfer and sintering) are combined to predict the porosity and bond quality.

\subsubsection{FFF temperature modeling}\label{sec:method1_1}
\begin{figure}
\centering
\includegraphics[width=0.65\textwidth]{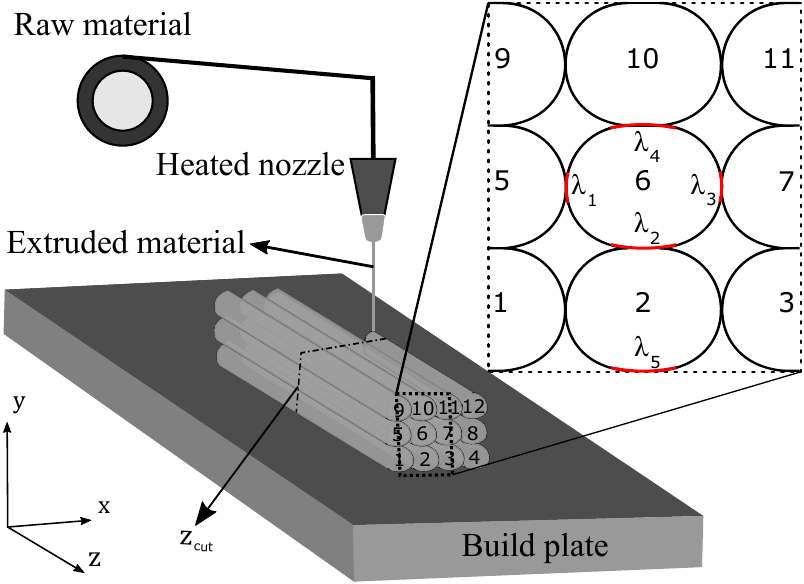}
\caption{Schematic of FFF.}
\label{fig:FFFschematic}
\end{figure}

Fused filament fabrication (FFF) (also known as fused deposition modeling or FDM\textsuperscript{\textregistered}) is an AM technology based on material extrusion for manufacturing polymer and plastic parts. The continuous strand of material is pushed through a heated nozzle and deposited as a molten extruded thin filament onto the build plate in a predefined path to form the part in the desired shape. The schematic of the FFF process is shown in Fig.~\ref{fig:FFFschematic}.

In FFF, each filament is subjected to the same heat transfer mechanism with different boundary conditions depending on thermal conditions such as environment, build plate and extrusion temperature, as well as the part geometry, material properties and deposition sequence. Recently, Costa et al.~\cite{costa2017estimation} developed an analytical solution for the transient heat transfer during the print process in FFF. The temperature prediction model considers conduction heat transfer with the build plate and adjacent filaments based on the fraction $\lambda_i$ of filament perimeter that is in contact, and convection heat transfer with the environment. The details of this heat transfer model are provided in the \emph{online supplementary material} (refer to online supplementary material).

\subsubsection{Bond formation modeling}\label{sec:method1_2}
This subsection develops a new method to compute the rate of polymer sintering and the final mesostructure of the printed part. The sintering process is defined as the coalescence of particles, in which two particles of molten polymer form a homogeneous melt, under the action of surface tension~\cite{pokluda1997modification}. The sintering process for amorphous polymers is driven by the surface tension force since the mechanism is considered a Newtonian viscous flow~\cite{rosenzweig1981sintering}.

A Newtonian sintering model for polymers was initially developed by Frenkel et al.~\cite{frenkel1945viscous} to predict the rate of polymer sintering. Pokluda et al.~\cite{pokluda1997modification} developed a closed-form equation to predict the bond formation between two spherical particles based on the work balance of viscous dissipation and surface tension. Bellehumeur et al.~\cite{bellehumeur2004modeling} applied the model proposed by Pokluda et al.~\cite{pokluda1997modification} to FFF for predicting the sintering between adjacent filaments as a nonlinear function of time, temperature-dependent surface tension $\Gamma(T)$ and viscosity $\eta(T)$, and an initial particle radius. The model has limitations related to the mesostructure of the filaments; specifically, the geometry of the filaments is assumed constant during the printing process. Based on the above discussion, in this paper we propose a new sintering model, which considers realistic filament geometry (similar to the one proposed by Garzon et al.~\cite{garzon2020design}) and also accounts for the change in the mesostructure of the filaments during the printing process.
\begin{figure}[!ht]
\centering
\includegraphics[width=0.8\textwidth]{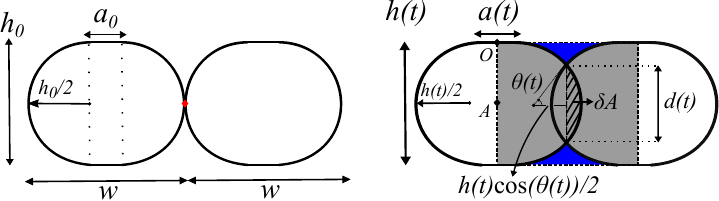}
\caption{Evolution of neck diameter during sintering process.}
\label{fig:Sintering}
\end{figure}

\begin{figure}[!ht]
\centering
\includegraphics[width=0.6\textwidth]{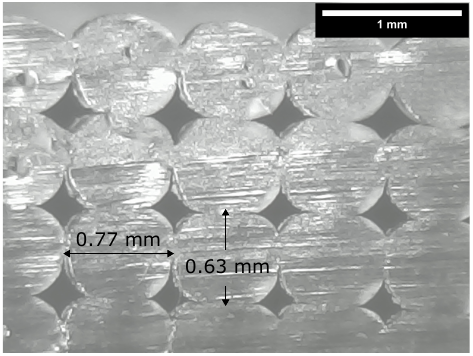}
\caption{Cross-sectional geometry of an FFF part printed with an extrusion temperature 240$^{\circ}$C, speed 42\ mm/s, initial layer height 0.7 mm, filament width 0.8 mm, filament length 35 mm, 6 number of layers and 15 filaments per layer.}
\label{fig:Experiment}
\end{figure}

The sintering process is simulated by considering two symmetrical adjacent filaments. In this work, all filaments are assumed to undergo the same sintering process by neglecting the effect of location. The cross-section geometry of a filament is composed of a rectangle with an initial width $a_0$ and two half circles with a radius of $h_0/2$ at initial time $t=0$ as proposed by Garzon et al.~\cite{garzon2020design}. At $t=0$, adjacent filaments have one contact point between them (see Fig.~\ref{fig:Sintering}). The width of the rectangle $a(t)$ evolves in time, together with $h(t)$ such that width of each filament $w = a_0 + h_0 = a(t) + h(t)\cos(\theta(t))$ stays constant. The layer height $h(t)$ is assumed to evolve in time based on the experimental observations, in which the average layer height for FFF parts has decreased more than the average width of filaments (see Fig.~\ref{fig:Experiment}). During the sintering process, the width and length of each filament ($w$ and $L$) are assumed to be constant. The coalescence between these two half circles forms the sintering angle $\theta(t)$ and neck diameter $d(t)$.

In order to calculate the change in the filament geometry during the sintering process, the law of conservation of mass is expressed for two adjacent filaments that are assumed to have constant density:
\begin{align}
    2L\Bigg(\pi \bigg(\frac{h_0}{2}\bigg)^2 + h_0a_0 \Bigg) 
    = 2L\Bigg(\pi \bigg(\frac{h(t)}{2}\bigg)^2 + h(t)a(t) - 2\delta A \Bigg),
\end{align}
where $2\delta A = 2(h(t)/2)^2(\theta(t)-\sin(2\theta(t)))$ is the area of intersection between two filaments as shown in Fig.~\ref{fig:Sintering}.

Thus, the evolution of layer height is obtained as
\begin{equation}
h(t) = \frac{-2 w + \sqrt{Q}}{\pi - 2 \theta + \sin{\left(2 \theta \right)} - 4 \cos{\left(\theta \right)}},
\end{equation}
where
\begin{align}
    Q  &= h_{0}^{2} \pi^{2} - 2 h_{0}^{2} \pi \theta + h_{0}^{2} \pi \sin{\left(2 \theta \right)} - 4 h_{0}^{2} \pi \cos{\left(\theta \right)} - 4 h_{0}^{2} \pi \notag\\ 
    &\ \ + 8  h_{0}^{2} \theta - 4 h_{0}^{2} \sin{\left(2\theta \right)} + 16 h_{0}^{2} \cos{\left(\theta \right)} + 4 h_{0} \pi w \notag\\
    &\ \ - 8 h_{0} \theta w + 4 h_{0} w \sin{\left(2 \theta \right)} - 16 h_{0} w \cos{\left(\theta \right)} + 4 w^{2}.
\end{align}

The work done by surface tension is defined as
\begin{equation}
    W_S = - \Gamma(T) \frac{dS}{dt},
\end{equation}
where $S$ is the total surface of the filaments that undergo sintering process and is given as
\begin{align}\label{surface}
    S(t) &= h(t)^{4} \left(- \theta + \frac{\sin{\left(2 \theta \right)}}{2}\right) + h(t)^{2} \pi + 2 h(t) L \pi \notag\\
    &\ \ - 4 h(t) L \theta + 4 h(t) a(t) + 4 L a(t).
\end{align}

Thus, applying the chain rule on Eq.~\eqref{surface}, the work done by surface tension is obtained as
\begin{equation}
    W_S = \Gamma(T) h(t) \left(2 h(t)^{3} \sin^{2}{\left(\theta \right)} + 4 L\right)\theta'.
\end{equation}

The work done by the viscous forces for a Newtonian fluid can be expressed as
\begin{equation}
    W_\nu = \int\int\int_V \eta(T)\nabla u : (\nabla u+\nabla u^{T}) dV,
\end{equation}
$V$ being the volume of the sintering system, and $\nabla u$ the gradient of velocity and is expressed as
\begin{equation}
    \nabla u = \begin{bmatrix}
                \dot{\epsilon_1} & 0 & 0\\
                0 & \dot{\epsilon_2} & 0\\
                0 & 0 & \dot{\epsilon_3}\\
                \end{bmatrix}.
\end{equation}
where $\dot{\epsilon_i}$ is the strain rate in $i$th direction.

Assuming the deformations in width and length are negligible w.r.t. deformations in height, $\dot{\epsilon_2}=\dot{\epsilon}$ is approximated by
\begin{equation}
    \dot{\epsilon} = \frac{\partial u_x(A)}{\partial x} \approx \frac{u_x(A)-u_x(O)}{\overline{OA}_0} = \frac{\frac{d}{dt} h(t)}{h_0}.
\end{equation}
Consequently, the work done by the viscous forces can be defined as follows
\begin{align}
    W_\nu &= \int\int\int_V 2\eta(T) \dot{\epsilon}^2 dV \\ &\quad = \frac{2 L \eta(T) h(t)^{2} \left(h_{0} \pi - 4 h_{0} + 4 w\right) \sin^{2}{\left(\theta \right)}}{h_{0}\cos^{2}{\left(\theta \right)}}\theta''.
\end{align}

The evolution of the sintering angle $\theta(t)$ is then obtained by equating the work done by surface tension and the viscous forces under the assumption that $\theta'$ is always positive~\cite{pokluda1997modification}:
\begin{equation}
  \frac{d\theta(t)}{dt} = \frac{\Gamma(T) h_{0} \left(h(t)^{3} + \frac{2 L}{\sin^{2}{\left(\theta(t) \right)}}\right)\cos^{2}{\left(\theta \right)}}{L h(t)\eta(T) \left(h_{0} \pi - 4 h_{0} + 4 w\right)}.
\end{equation}
The evolution of neck diameter $d(t)$ with time is then computed as
\begin{equation}
    d(t) = h(t) \sin(\theta(t)),
\end{equation}

The porosity can be expressed using the geometry of the mesostructure shown in Fig.~\ref{fig:Sintering}, where the shaded grey region is the filled area and blue region is the void area. Thus, the evolution of porosity $\phi(t)$ is calculated as
\begin{equation}
    \phi(t) = \frac{ h^{2} \cos{\left(\theta \right)} - \left(\frac{h}{2}\right)^2 \big[\pi - \left(2\theta - \sin{\left(2\theta \right)}\right)\big]}{h^{2} \cos{\left(\theta \right)} + h a}.
\end{equation}

The proposed methodology is an improvement upon the previous work on sintering models by Pokluda et al.~\cite{pokluda1997modification} and Gurrala et al.~\cite{gurrala2014part}, because (i) it considers a realistic filament geometry based on our experiments, and (ii) it accounts for changes in the filament geometry during the printing process. This proposed methodology is compared against experimental data in Section~\ref{sec:results}, and is found to give a smaller error than Gurrala et al.'s model. However, it still makes assumptions about the bonding process and sintering is not the only physical phenomenon that takes place during the bond formation. Thus, deep learning is studied to further enhance the prediction accuracy.

\subsection{PIML for additive manufacturing}\label{sec:method3}
Although physics-based models predicting the temperature evolution, bond formation and mesostructure evolution of FFF parts are based on physical laws, they introduce bias due to incomplete representation of the complex physical process by approximating the reality. In addition, these models contain a significant number of model parameters that need to be calibrated using experimental data~\cite{berkcan}. On the other hand, ML models are not aware of physical laws, which may result in physically inconsistent model predictions. However, they can extract complex physical relationships from available data. Thus, physics-based models and ML models can be integrated in an innovative manner to better capture the dynamics of the AM process.

In PIML models, physics knowledge and data are sought to be integrated in a synergistic manner by leveraging the complementary strengths of both models~\cite{karpatne2017physics}. Thus, the goal is to improve the predictions beyond that of physics-based models or ML models alone by coupling physics-based models with ML models. In the following, three different strategies to combine physics knowledge and ML models are pursued: (1) incorporate physics constraints within the loss function of the DNN, (2) use physics model outputs as additional inputs to the DNN model, and (3) pre-train a DNN model with physics model input-output and then update it with experimental data.

\subsubsection{Physics-informed loss functions}
A direct strategy to improve ML model predictions is by including physics-based loss functions~\cite{karpatne2017physics}. Consider a PIML model with inputs $\boldsymbol{X}$ and outputs $\boldsymbol{\hat{Y}}$ trained using physical laws that are incorporated as constraints into the loss function:
\begin{align}\label{eq:lossfunc}
    \mathcal{L} = \mathcal{L}_{\rm DNN}(\boldsymbol{Y}, \boldsymbol{\hat{Y}}) + \sum_{k=1}^M \lambda_{\rm phy, k}\mathcal{L}_{{\rm phy}, k}(\boldsymbol{\hat{Y}}),
\end{align}
where $\mathcal{L}_{\rm DNN}$ is the regular training loss of a DNN that evaluates a supervised error (e.g., root mean squared error (RMSE); $\mathcal{L}_{\rm DNN}(\boldsymbol{Y}, \boldsymbol{\hat{Y}})=\sqrt{\sum_{i=1}^{n} (Y_{i}- \hat{Y}_{i})^2/n}$), which measures how far off the predictions $\boldsymbol{\hat{Y}}$ are from the observations $\boldsymbol{Y}$ for the $n$ training samples, and $\mathcal{L}_{\rm phy,k}$ is the $k$-th physics-based loss function, whose contribution is controlled by a hyperparameter $\lambda_{\rm phy,k}$ and $M$ is the total number of physics-based loss functions. The inclusion of $\mathcal{L}_{\rm phy,k}$ ensures physically consistent model predictions (the second term of Eq.{~\eqref{eq:lossfunc}} means \emph{physical inconsistency}) and can decrease the generalization error even when there is a small amount of training data~\cite{karpatne2017physics}. In addition, $\mathcal{L}_{\rm phy,k}$ does not require experimental observations; the data obtained from the physics model is used to evaluate physics-based loss functions.

In this work, we enforce five different physics-based loss functions (i.e., five separate physical relationships, $\mathcal{L}_{{\rm phy}, k}(\boldsymbol{\hat{Y}})$, where $k=\{1,2,3,4,5\}$ and $\boldsymbol{\hat{Y}}=(\hat{Y}_{1}, \hat{Y}_{2})$ are the overall dimensionless neck diameter (i.e., overall bond quality) and porosity predictions of FFF parts, respectively). These loss functions are defined as follows:
\begin{align}
    & {}\mathcal{L}_{{\rm phy}, 1}(\boldsymbol{\hat{Y}}) = \frac{1}{N} \sum_{i=1}^N \rm{ReLU}(-\mathnormal{\hat{Y}}_{1,i}), \notag\\
    & {}\mathcal{L}_{{\rm{phy}}, 2}(\boldsymbol{\hat{Y}}) = \frac{1}{N} \sum_{i=1}^N \rm{ReLU}(\mathnormal{\hat{Y}}_{1,i}-\mathnormal{d_{max}}), \notag\\
    & {}\mathcal{L}_{{\rm phy}, 3}(\boldsymbol{\hat{Y}}) = \frac{1}{N} \sum_{i=1}^N \rm{ReLU}(-\mathnormal{\hat{Y}}_{2,i}), \notag\\
    & {}\mathcal{L}_{{\rm phy}, 4}(\boldsymbol{\hat{Y}}) = \frac{1}{N} \sum_{i=1}^N \rm{ReLU}(\mathnormal{\hat{Y}}_{2,i}-\phi_{0,i}), \notag\\
    & {}\mathcal{L}_{{\rm phy}, 5}(\boldsymbol{\hat{Y}}) = \frac{1}{N} \sum_{i=1}^N \rm{ReLU}(\Delta_i),
\end{align}
where the first four loss functions consider the physical violations related to the overall dimensionless neck diameter and porosity across $N$ samples and the fifth loss function represents the physical relationship between the mechanical properties and neck diameter. The physical inconsistencies in the model predictions are evaluated using these physics-based loss functions. In the first and third loss functions, negative values of neck diameter and porosity are treated as physical violations. The second loss function evaluates physically inconsistent dimensionless neck diameter predictions which are greater than the maximum dimensionless neck diameter $d_{max}=1$. The fourth loss function penalizes the model when porosity predictions $\mathnormal{\hat{Y}}_{2,i}$ are greater than the initial porosity $\phi_{0,i}$ of $i$th part. This is based on the physics knowledge that the total void area decreases as the sintering process takes place. The fifth physics-based loss function exploits the monotonic relationship between bond quality and tensile strength of FFF-produced parts. This loss function is constructed by computing the difference in the sorted dimensionless neck diameter predictions, $\mathnormal{\hat{Y}}_{1,\rm{sorted}}$, and dimensionless neck diameter predictions corresponding to sorted tensile strength estimates ($\sigma_{TS}(\mathnormal{\hat{Y}}_{2,\rm{sorted}}, \boldsymbol{\xi})$), $\mathnormal{\hat{Y}^{'}}_{1,i}$, i.e., $\Delta_i = \mathnormal{\hat{Y}}_{1,\rm{sorted},i} - \mathnormal{\hat{Y}^{'}}_{1,i}$. The maximum stress for longitudinal raster orientation, $\sigma_{TS}(\mathnormal{\hat{Y}}_{2}, \boldsymbol{\xi})$ with $\boldsymbol{\xi}=\{\sigma_{01}, \sigma_{02}, C_{\sigma}\}$ being the material parameters, is computed according to the analytical expression proposed in Garzon et al.~\cite{garzon2020design} using the porosity predictions:
\begin{align}
    \sigma_{TS} = \sigma_{01} \bigg[\rm{exp}\Big( (1-\mathnormal{\hat{Y}}_{2})^{C_{\sigma} n_l } \Big) - \mathnormal{\hat{Y}}_{2} \bigg] + \sigma_{02}(1-\mathnormal{\hat{Y}}_{2}),
\end{align}
where $n_l$ is the number of layers of FFF parts. The tensile strength is constrained when porosity is equal to 0, i.e., $\sigma_{TS}=\sigma_{01}e + \sigma_{02}$. The overall average neck diameter and tensile strength are positively correlated, and tensile strength increases monotonically with neck diameter. Whereas, porosity and tensile strength are negatively correlated (as are porosity and neck diameter). More specifically, the model predictions $(\hat{Y}_{1,i}, \hat{Y}_{2,i})$ and $(\hat{Y}_{1,i+1}, \hat{Y}_{2,i+1})$ corresponding to $i$th and $(i+1)$th FFF parts can be used to estimate $\sigma_{TS,i}$ and $\sigma_{TS,i+1}$. If $\sigma_{TS,i+1}$ is greater than $\sigma_{TS,i}$ --- meaning $(i+1)$th part has less voids than $i$th part ($\hat{Y}_{2,i+1}<\hat{Y}_{2,i}$)---then $\hat{Y}_{1,i+1}$ should be greater than $\hat{Y}_{1,i}$ as well by exploiting a key monotonic physical relationship between porosity and tensile strength of FFF-produced parts. Thus, with the inclusion of these physics-based penalty functions, the neck diameter and porosity predictions are ensured to be physically meaningful.

\subsubsection{Physics model output as additional ML model input}
A physics-based model $f^{\rm phy}: \boldsymbol{X} \rightarrow \boldsymbol{\hat{Y}}^{\rm phy}$ can be used to predict the QoI, where $\boldsymbol{\hat{Y}}^{\rm phy}$ are predicted estimates of the true response of the system $\boldsymbol{Y}$. A straightforward approach to combine physics-based and ML models is to use physics model output $\boldsymbol{\hat{Y}}^{\rm phy}$ (at the experimental inputs) as additional input along with inputs $\boldsymbol{X}$; i.e., $f^{\rm hyb}: \boldsymbol{X}^{\rm hyb}=[\boldsymbol{X}, \boldsymbol{\hat{Y}}^{\rm phy}] \rightarrow \boldsymbol{\hat{Y}}^{\rm hyb}$.

Adding the physics output as an extra input to the DNN model (which is trained using experimental data) is information fusion, where the physics model is an additional source of information that is consistent with physics (i.e., when the physics model satisfies the constraints mentioned in the previous subsection). The resulting DNN model can be thought of as a hybrid model that uses the experimental data to correct the output of the physics model which is an incomplete representation of the actual physics.

\subsubsection{Pre-trained PIML model}\label{sec:method4}
In AM, especially in the FFF process with not a high-quality printer, parts have significant variability in quality. There is also uncertainty in measurement and lack of data due to the high cost associated with conducting experiments. Thus, data of adequate quality and quantity is important for good quality model predictions in AM.

In order to leverage the complex physical knowledge inherent in the physics-based models, synthetic data can be generated for multiple input combinations using physics-based models. The synthetic data can be used to train a ML model, which is used as the initial model to be updated with experimental data. The transfer of physical knowledge using a pre-trained ML model can prevent poor initialization due to lack of knowledge of initial choice of ML model parameters prior to training. This allows the pre-trained ML model to be fine-tuned even with limited observed data. In addition, it has been shown that using synthetic data from even imperfect physics models with uncalibrated model parameters can still reduce the amount of experimental training data needed~\cite{jia2020physics}.

More importantly, the pre-training can use a large amount of training data (with multiple input parameter combinations) over a wide range of values, which is not possible in experiments that could be expensive; as a result, the pre-training may help the eventual ML model to have wider generalization beyond experimental data. This is also an important distinction of the pre-training strategy from the second strategy. Both strategies use the physics model, but in the second strategy, the physics model is only used to provide outputs corresponding to the experimental inputs, whereas in the current pre-training strategy, the physics model is used to provide outputs corresponding to a much larger set of inputs. In the numerical example in Section~\ref{sec:results}, the pre-training strategy exercises the physics model over 1525 input combinations, whereas the second strategy above only employs the physics model over 39 experimental input combinations. However, the advantage of the pre-training strategy in using a larger input data set (for physics model runs) compared to the experiments becomes limited if the physics model is computationally expensive.

In this work, the ML model is pre-trained using the outputs of an uncalibrated coupled multi-physics model (i.e., neck diameter and porosity). Further, the transfer of learned physical knowledge is shown to be valuable even when the input parameters of the synthetic data generated are quite different than the experimental observations. Once the ML model is pre-trained, it is fine-tuned using limited experimental observations. This helps to learn a 3D printer-specific physical process faster and with less samples.

\begin{figure}[!ht]
	\centering
	\includegraphics[width=0.8\textwidth]{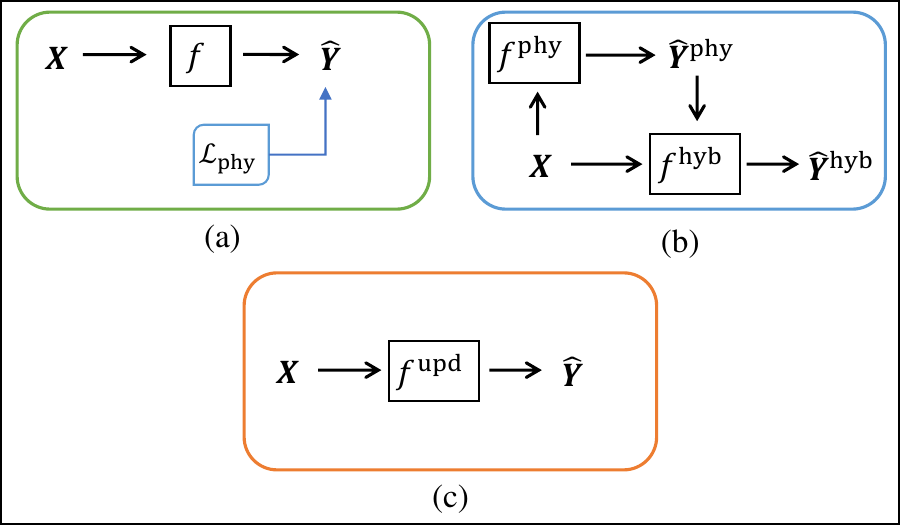}
	\caption{PIML strategies: (a) incorporate physics constraints within the loss function of the DNN, (b) use physics model outputs as additional inputs to the DNN model, and (c) pre-training a DNN model with physics model input-output and updating it with experimental data.}
	\label{fig:diagram}
\end{figure}

The three proposed strategies to predict the QoIs are shown in Fig.~\ref{fig:diagram}. Figure~\ref{fig:diagram}(a) shows the first method, where the physical knowledge is included through constraints within the loss function of a DNN trained with experimental data. Figure~\ref{fig:diagram}(b) shows the second method, where the outputs of the physics model are additional inputs to the DNN model. Figure~\ref{fig:diagram}(c) shows the third method, where a DNN model is pre-trained with data generated using the physics-based model and then updated using experimental data. The proposed PIML strategies can be applied to any AM process by leveraging the physical constraints or physics-based models.

\subsubsection{Combination of PIML strategies}\label{sec:method5}
Based on the proposed three strategies to incorporate physics knowledge into the ML model, eight separate ML models can be constructed:
\begin{multicols}{2}
    \begin{enumerate}
        \item $\mathbf{\rm DNN}$
        \item $\mathbf{\rm DNN^{\mathcal{L}_{\rm phy}}}$
        \item $\mathbf{\rm DNN^{\rm hyb}}$
        \item $\mathbf{\rm DNN^{\rm upd}}$
        \item $\mathbf{\rm DNN^{\rm hyb, \mathcal{L}_{\rm phy}}}$
        \item $\mathbf{\rm DNN^{\rm upd, hyb}}$
        \item $\mathbf{\rm DNN^{\rm upd, \mathcal{L}_{\rm phy}}}$
        \item $\mathbf{\rm DNN^{\rm upd, hyb, \mathcal{L}_{\rm phy}}}$
    \end{enumerate}
\end{multicols}

In model 1, a deep neural network $\mathbf{\rm DNN}$ is trained using only experimental data. The inputs $\boldsymbol{X}$ for this basic DNN model are the process parameters, printer extrusion temperature, speed, layer height, filament width, length, number of layers, and number of filaments per layer; and the outputs are overall dimensionless neck diameter and porosity. These inputs and outputs are the same as those used in the physics-based model $f^{\rm phy}$. Model 2 ($\mathbf{\rm DNN^{\mathcal{L}_{\rm phy}}}$) pursues the first strategy: physical knowledge related to the FFF process is included through constraints within the loss function of the DNN as shown in Eq.~\eqref{eq:lossfunc}. Model 3 pursues the second strategy: a hybrid physics-based neural network $\mathbf{\rm DNN^{\rm hyb}}$ is trained using the outputs $\boldsymbol{\hat{Y}}^{\rm phy}$ of $f^{\rm phy}$ as extra inputs in addition to $\boldsymbol{X}$, i.e., $\boldsymbol{X}^{\rm hyb}=[\boldsymbol{X}, \boldsymbol{\hat{Y}}^{\rm phy}]$. Model 4 ($\mathbf{\rm DNN^{\rm upd}}$) pursues the third strategy, where the weights and biases (model parameters) of all the layers \emph{excluding} the input layer of the pre-trained network $f^{\rm pre}$ (which is trained with the coupled multi-physics model input-output described in Section~\ref{sec:method4}) are used as initial parameters for the DNN model and these parameters are updated with experimental data. 

The architecture of the pre-trained model and the updated models (Model 4, 6, 7, and 8) is the same (except the input layer which changes with different numbers of inputs). The rest of the models (5-8) represent the combinations of the three strategies. Models 5, 6, and 7 each combine any two of the three strategies, whereas model 8 combines all three strategies. Model 5 ($\mathbf{\rm DNN^{\rm hyb, \mathcal{L}_{\rm phy}}}$) combines the use of physics model outputs as additional inputs and the incorporation of physics constraints $\mathcal{L}_{\rm phy}$ within the loss function of the DNN. Model 6 ($\mathbf{\rm DNN^{\rm upd, hyb}}$) combines second and third strategies, where the optimized model parameters of $f^{\rm pre}$ are used as the initial values and the outputs $\boldsymbol{\hat{Y}}^{\rm phy}$ of $f^{\rm phy}$ are included as additional inputs. $\mathbf{\rm DNN^{\rm upd, hyb}}$ has the same number of inputs as $\mathbf{\rm DNN^{\rm hyb}}$ (i.e., both include the physics model output as additional input) and uses the optimized model parameters of all the layers excluding the input layer of $f^{\rm pre}$ as initial parameters before updating with experimental data. Model 7 ($\mathbf{\rm DNN^{\rm upd, \mathcal{L}_{\rm phy}}}$) combines first and third strategies. The model parameters obtained from $f^{\rm pre}$ are updated using the experimental data by minimizing the augmented loss function shown in Eq.~{\ref{eq:lossfunc}}. Model 8 ($\mathbf{\rm DNN^{\rm upd, hyb, \mathcal{L}_{\rm phy}}}$) combines the use of physics model outputs as additional inputs to the updated DNN model $\mathbf{\rm DNN^{\rm upd}}$ (which results in $\mathbf{\rm DNN^{\rm upd, hyb}}$) and the physics constraints $\mathcal{L}_{\rm phy}$ are incorporated within the loss function of $\mathbf{\rm DNN^{\rm upd, hyb}}$.

\section{Implementation of PIML to FFF}\label{sec:results}
In this section we demonstrate the implementation of the proposed methodology to FFF-produced parts, and investigate the performance of eight PIML models described in Section~\ref{sec:method5}. The results show that the proposed PIML models are capable of achieving physically meaningful and accurate model predictions, and require a smaller number of experiments.

\subsection{Problem setup}\label{sec:results1}
First, data is collected from laboratory experiments in order to build the prediction models for the QoI using ML techniques. The shape of the part is conceptualized, a CAD model is visualized, and then sliced in a slicing software using the defined FFF process parameters and printing path.

The print quality depends on the adhesion of the first layer with the build plate~\cite{berkcan}. Thus, several measures are needed to ensure proper adhesion. For example, the printing environment is modified by adding an enclosure to the 3D printer to isolate the printing environment from external effects. Kapton tape is used on the glass build plate to enhance the adhesion of Acrylonitrile butadiene styrene (ABS) with the build plate. After these modifications, the part is printed and then measured with appropriate monitoring techniques.

A commercial material Ultimaker Black ABS was extruded from an Ultimaker 2 Extended+ 3D printer to manufacture parts with unidirectionally aligned filaments. Using Latin hypercube sampling, 20 sets of process parameters are generated. The ranges considered for the variables are printer extrusion temperature $T_e$: (210$^{\circ}$C - $260 ^{\circ}$C), and extrusion speed $S_e$: (15 mm/s - 46 mm/s). Since the values of material properties $\boldsymbol{\xi}$ do not affect the outcome of $\mathcal{L}_{{\rm phy}, 5}(\boldsymbol{\hat{Y}})$, the values calibrated by Garzon et al.~\cite{garzon2020design} are used. All specimens were sectioned at the midpoint $z_{\rm cut}=L/2$ (since the statistical properties of QoIs along the length of the specimens were constant) to analyze the mesostructural feature of interest with the use of microscopy images processed through the ImageJ software~\cite{schneider2012nih}. The collected experimental data is subsequently used to create DNN prediction models.

\subsection{Model training and prediction}
\begin{figure}[!ht]
	\centering
	\includegraphics[width=0.6\textwidth]{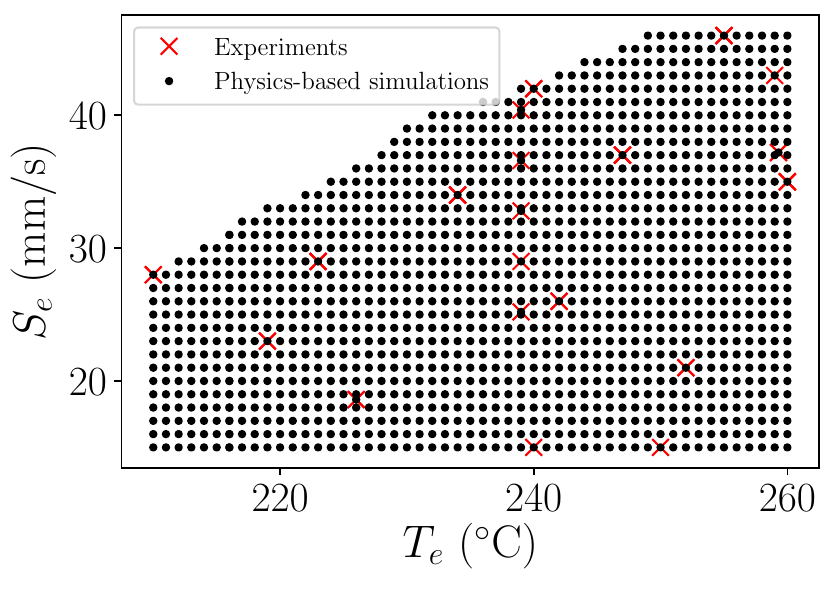}
	\caption{Design of experiments for physics-based simulations and experiments.}
	\label{fig:DoE}
\end{figure}

The eight DNN models were implemented using the Keras package~\cite{chollet2015keras} with Tensorflow backend. The pre-trained model, $f^{\rm pre}$, is first trained with physics model input-output data consisting of 1525 input parameter combinations over a range of experimental values, i.e., (210$^{\circ}C$  $\le T_e \le$ $260 ^{\circ}C$,  15 mm/s $\le S_e\le$ 46 mm/s ), and then updated for different combinations of the proposed strategies using observed data. (Note that in contrast only 39 physical experiments with 20 unique input parameter combinations are available, see Fig.~\ref{fig:DoE}). The input data of the training and test sets are normalized prior to the training of the DNN models (the output quantities are dimensionless and between 0 and 1), and the hyperparameters of these models are tuned with grid search ($\lambda_{\rm phy}= 0.3, 0.3, 0.15, 0.15, 0.008$). Fully-connected DNN models with 2 hidden layers and 10 neurons in each hidden layer are constructed and the weights of all neurons in these models are uniformly randomly initialized between 0 and 1. L1 and L2 regularizers are used as a penalty term to avoid overfitting. The Rectified Linear Unit (ReLU) activation function and Adam optimizer are used to perform stochastic gradient descent in learning the model parameters.

The number of epochs for the convergence of training is approximately the same for each model except $f^{\rm pre}$, which converges in 40 epochs. The computation time for training of each model is on average 15 sec using a desktop computer (Intel\textsuperscript{\tiny\textregistered} Xeon\textsuperscript{\tiny\textregistered} CPU E5-1660 v4$@$3.20GHz with 32 GB RAM and GPU NVIDIA Quadro K620 with 2 GB).

\subsection{Model performance}
In order to measure the accuracy of the trained DNN models, the model predictions of the test data are compared with the observed overall neck diameters and porosity. Each model is trained 30 times and compared against the FFF experimental data not used for training (i.e., data from 19 parts) to evaluate the mean and standard deviation of RMSE and physical inconsistency. The effect of the training data size on the RMSE of different models is shown in Table~\ref{table:EffectOfTrainingData}. Here, the models are trained with 4, 6, 8, 10, and 20 experimental data points (i.e., 20\%, 30\%, 40\%, 50\%, and 100\% of the available training data). The RMSE values of the coupled-physics model based on the sintering model developed by Gurrala et al.~\cite{gurrala2014part} and the new proposed one are 0.173 and 0.112, respectively, which are larger than the ML models due to the approximations used to represent the FFF process and bias in the model. The mean RMSE value of $f^{\rm pre}$, which is trained with physics model input-output data consisting of 1525 input parameter combinations is 0.362. The RMSE value of $f^{\rm pre}$ is greater than the RMSE of $f^{\rm phy}$ because the pre-trained model is an approximation of the physics model, which causes uncertainty and bias. The RMSE of the basic DNN model is about 0.025 when we use all the parts in the training set. The RMSE of the basic DNN model is better than $f^{\rm pre}$ and $f^{\rm phy}$ because experimental data is directly used as the training data for the basic DNN model, whereas $f^{\rm pre}$ and $f^{\rm phy}$ use the approximate physics model to generate either pre-training data or additional input to the ML model.

The physics-based loss functions allow $\mathbf{\rm DNN^{\mathcal{L}_{\rm phy}}}$ and $\mathbf{\rm DNN^{\rm hyb, \mathcal{L}_{\rm phy}}}$ to achieve physically meaningful results with a lower value of average RMSE than $\mathbf{\rm DNN^{\rm hyb}}$ and improve the generalization performance. The $\mathbf{\rm DNN^{\rm hyb}}$ model performs similar to $\mathbf{\rm DNN}$, which shows that the physics model outputs do not improve the learning process significantly. The $\mathbf{\rm DNN^{\rm upd}}$ and $\mathbf{\rm DNN^{\rm upd, hyb}}$ models achieve a similar performance improvement as the DNN models that include physics-based loss functions. However, the models without physics constraints produce physically inconsistent results as shown in Table~\ref{table:EffectOfTrainingData}. The results show that pre-training the PIML model improves the performance, and the improvement is relatively larger as the amount of observed data gets smaller. Additionally, the models that are pre-trained ($\mathbf{\rm DNN^{\rm upd}}$ and $\mathbf{\rm DNN^{\rm upd, hyb}}$) reach a physically more consistent initialization when they are updated with experimental data even without using physics constraints, compared to models that are not pre-trained, $\mathbf{\rm DNN}$ and $\mathbf{\rm DNN^{\rm hyb}}$. The combination of all strategies ($\mathbf{\rm DNN^{\rm upd, hyb, \mathcal{L}_{\rm phy}}}$) allows the model to get closest to the ground truth.
\begin{table}
    \caption{Effect of different amounts of training data on the RMSE of different ML models}
    \label{table:EffectOfTrainingData}
    \resizebox{1\columnwidth}{!}{%
        \begin{tabular}{@{\extracolsep{\fill}}l*{5}c@{\extracolsep{\fill}}  c}
        \hline\noalign{\smallskip}
             & & & & & & Mean Physical\\
            Model& 20\% & 30\% & 40\% & 50\% & 100\% & Inconsistency\\
        \noalign{\smallskip}\hline\noalign{\smallskip}
            $f^{\rm phy}$ & - & - & - & - & 0.112 & \textbf{0.000}\\
            1. $\mathbf{\rm DNN}$ & 0.101($\pm$0.022) & 0.084($\pm$0.018) & 0.044($\pm$0.017) & 0.028($\pm$0.004) & 0.025($\pm$0.005) & 0.201\\
            2. $\mathbf{\rm DNN^{\mathcal{L}_{\rm phy}}}$ & 0.055($\pm$0.017) & 0.050($\pm$0.011) & 0.025($\pm$0.005) & 0.024($\pm$0.007) & 0.021($\pm$0.005) & \textbf{0.000}\\
            3. $\mathbf{\rm DNN^{\rm hyb}}$ & 0.098($\pm$0.021) & 0.078($\pm$0.024) & 0.044($\pm$0.016) & 0.028($\pm$0.007) & 0.024($\pm$0.005) & 0.195\\
            4. $\mathbf{\rm DNN^{\rm upd}}$ & 0.058($\pm$0.012) & 0.049($\pm$0.011) & 0.026($\pm$0.005) & 0.025($\pm$0.005) & 0.020($\pm$0.005) & 0.133\\
            5. $\mathbf{\rm DNN^{\rm hyb, \mathcal{L}_{\rm phy}}}$ & 0.059($\pm$0.018) & 0.041($\pm$0.012) & 0.026($\pm$0.004) & 0.023($\pm$0.004) & 0.020($\pm$0.005) & \textbf{0.000}\\
            6. $\mathbf{\rm DNN^{\rm upd, hyb}}$ & 0.057($\pm$0.014) & 0.047($\pm$0.011) & 0.026($\pm$0.003) & 0.026($\pm$0.005) & 0.024($\pm$0.006) & 0.134\\
            7. $\mathbf{\rm DNN^{\rm upd, \mathcal{L}_{\rm phy}}}$ & 0.054($\pm$0.011) & 0.045($\pm$0.013) & 0.026($\pm$0.003) & 0.025($\pm$0.005) & 0.021($\pm$0.005) & \textbf{0.000}\\
            8. $\mathbf{\rm DNN^{\rm upd, hyb, \mathcal{L}_{\rm phy}}}$ & 0.053($\pm$0.013) & 0.040($\pm$0.010) & 0.025($\pm$0.004) & 0.023($\pm$0.004) & 0.018($\pm$0.003) & \textbf{0.000}\\
        \noalign{\smallskip}\hline
        \end{tabular}
    }%
\end{table}

\begin{figure}[!ht]
	\centering
	\includegraphics[width=0.6\textwidth]{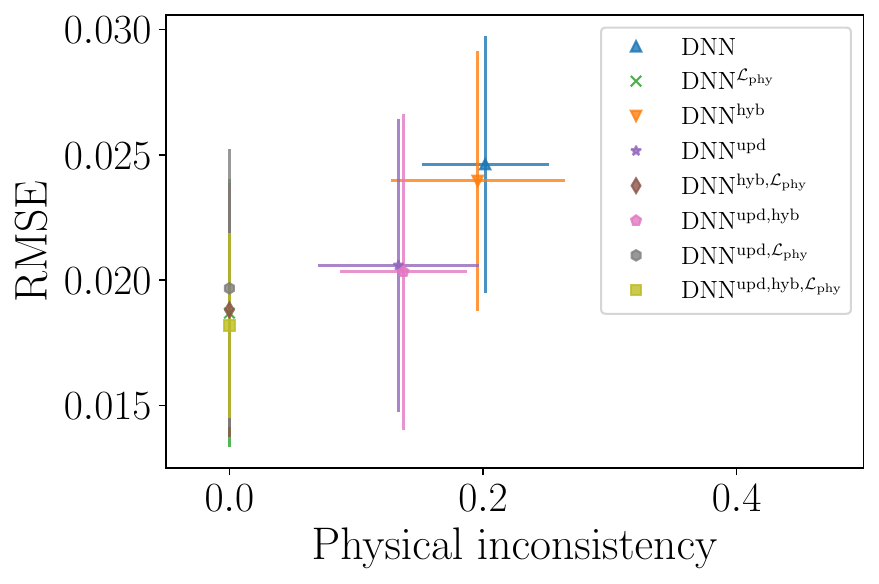}
	\caption{Performance and physical inconsistency of proposed models.}
	\label{fig:Performance}
\end{figure}

The prediction accuracy (w.r.t. neck diameter and porosity) of different models trained with 100\% of the experimental data is shown in Fig.~\ref{fig:Performance}. The x and y-axis represent the physical inconsistency and the mean and standard deviation of RMSE, respectively. Figure~\ref{fig:Performance} shows that models with physics-based loss functions produce physically consistent results. The incorporation of the physics knowledge using either the first or third strategy enables the models 2, 4-8 to generalize to configurations unseen in the training set.

\begin{figure}[!ht]
	\centering
	\includegraphics[width=1.0\textwidth]{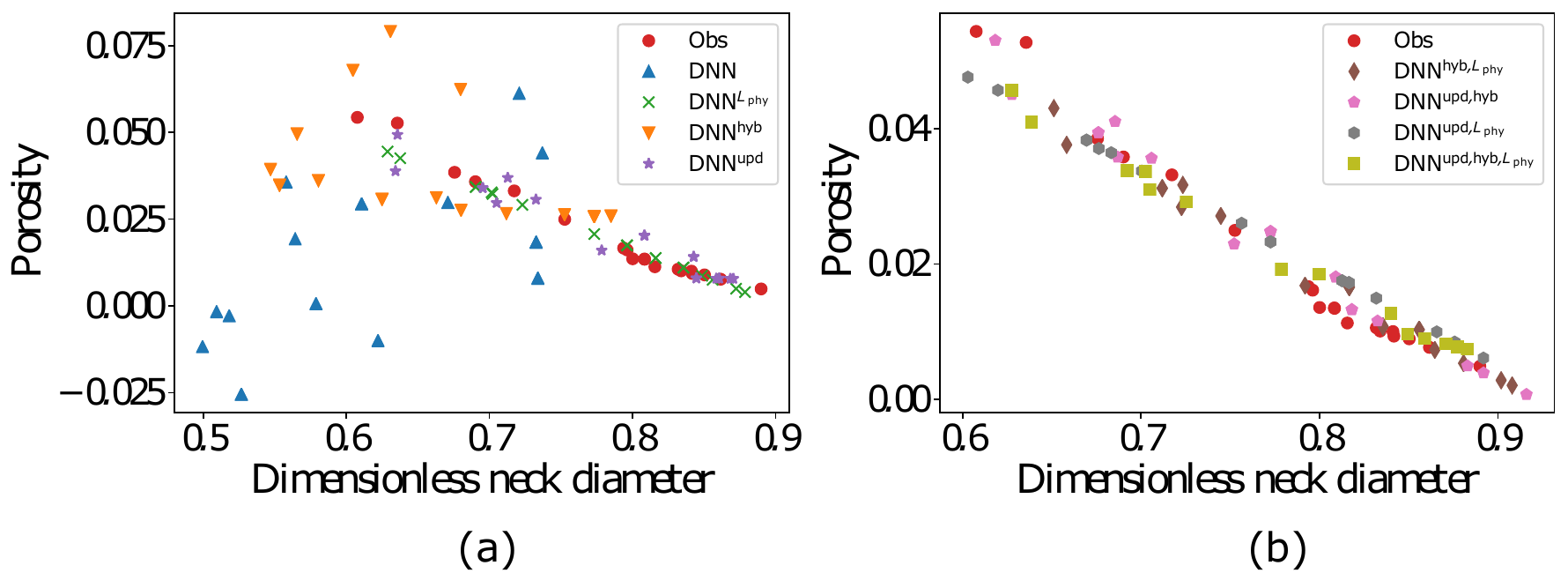}
	\caption{Comparison of model prediction with test data:~(a) for the first 4 models,~and (b) for the last 4 models.}
	\label{fig:Analysis_1and2}
\end{figure}

In order to further analyze the improvement in model predictions, the predicted dimensionless overall neck diameter and porosity for the test set that comprises 19 FFF parts are visualized in Fig.~\ref{fig:Analysis_1and2}. Porosity predictions of model 1 ($\mathbf{\rm DNN}$) have large physical inconsistency (Fig.~\ref{fig:Analysis_1and2}(a)). For instance, the blue triangle with the lowest porosity prediction has a negative value (i.e., $(\hat{Y}_{1}, \hat{Y}_{2}) = (0.52, -0.025)$). Model 2 ($\mathbf{\rm DNN^{\mathcal{L}_{\rm phy}}}$) predictions are physically consistent due to enforced physics constraints, but model 3 ($\mathbf{\rm DNN^{\rm hyb}}$) predictions also do not follow a monotonic decreasing relationship. Model 4 and 6 ($\mathbf{\rm DNN^{\rm upd}}$ and $\mathbf{\rm DNN^{\rm upd, hyb}}$) have some physically inconsistent predictions. Figure~\ref{fig:Analysis_1and2}(b) shows that $\mathbf{\rm DNN^{\rm hyb, \mathcal{L}_{\rm phy}}}$, $\mathbf{\rm DNN^{\rm upd, \mathcal{L}_{\rm phy}}}$ and $\mathbf{\rm DNN^{\rm upd, hyb, \mathcal{L}_{\rm phy}}}$ produce physically consistent model predictions, i.e., porosity and neck diameter predictions follow a monotonically decreasing relationship.

\section{Conclusion}\label{sec:conclusion}
In this work, three strategies for physics-informed machine learning (PIML) are investigated for predicting the quality-related metrics of FFF-produced parts. First, a physics-based sintering model is developed to predict the overall average neck diameter and porosity of FFF parts, using the temperature evolution of filaments, material properties, part geometry, and process parameters as inputs. The developed sintering model offers two improvements over existing models: (i) consideration of realistic filament geometry, and (ii) allowing the filament geometry to change during the printing process. Next, several PIML models are developed to predict the bond quality and porosity of FFF parts by leveraging three strategies for incorporating physics knowledge into the DNN model: (1) physics-based loss functions, (2) using the outputs of the coupled multi-physics model as additional inputs to the DNN, and (3) pre-training a DNN with data generated using physics-based model and then updating it with experimental data. The physics-based loss functions exploit the relationship between bond quality and tensile strength of the FFF parts.

The numerical results show that the incorporation of physics knowledge not only improves the prediction accuracy while producing physically meaningful results, but also allows accurate model predictions even with smaller amounts of experimental data. Thus, the proposed approach helps to fill the physics knowledge gap in the ML model while leveraging the capability of ML to extract complex process-material-geometry relationships in AM, and correcting for the approximation in the physics-based model.

Future work can include experimental data that consists of higher dimensional input, with multiple combinations, in order to further evaluate the performance of the proposed PIML strategies. Future work can also explore the generalization capabilities of the proposed strategies to parts of different geometry, as well as transfer learning to parts manufactured with different 3D printers and materials. Also, in the context of the third strategy, the data produced by experiments and physics models have different levels of credibility; thus the weighting of the two sources of data needs to be investigated in the future.

\section*{Conflict of Interest}
On behalf of all authors, the corresponding author states that there is no conflict of interest.

\bibliographystyle{unsrt}      
\bibliography{references}   

\end{document}